\documentclass[a4paper,11pt]{esslli}
\usepackage[leqno,tbtags]{amsmath2000}
\usepackage{amsthm2000}
\usepackage{gb4e}
\exewidth{(9.99)}
\usepackage[longnamesfirst]{natbib}

\offprhead{Proceedings of the Seventh ESSLLI Student Session}
\offprocc{\textnormal{Malvina Nissim (editor)}}
\offprauthors{Chung-chieh Shan and Balder ten Cate}

\begin{document}

\theoremstyle{plain}
\newtheorem{SHANtheorem}{Theorem}
\newtheorem{SHANcorollary}{Corollary}
\theoremstyle{definition}
\newtheorem{SHANdefinition}{Definition}
\newcommand{\SHANliff}{\leftrightarrow}
\newcommand{\SHANq}{\mathopen?}
\newcommand{\SHANapprox}{\mathbin\approx}
\newcommand{\SHANphrase}[1]{\emph{\frenchspacing#1}}
\newcommand{\SHANwh}{\SHANphrase{wh}\nobreakdash}
\newcommand{\SHANfun}[1]{\mathopen{\lambda\mathord{#1}.\,}}

\makeatletter
\let\c@exx\c@equation
\def\thexnumi{\thechapter.\@xsi{xnumi}}
\makeatother

\newcommand{\SHANcitearound}[2]{{\let\realspace\ \def\ {\let\ \realspace#1}{#2}}}
\newcommand{\SHANcitegenitive}{\SHANcitearound{'s }}

\chapter{The Partition Semantics of Questions, Syntactically}
\thispagestyle{firstpage}
\chapterauthor{Chung-chieh Shan}
\chapteraffiliation{Harvard University}
\chapteremail{ccshan@post.harvard.edu}

\chapterauthor{Balder ten Cate}
\chapteraffiliation{Universiteit van Amsterdam}
\chapteremail{b.ten.cate@hum.uva.nl}

\begin{chapterabstract}
    Groenendijk and Stokhof
    \citetext{\citeyear{groenendijk-studies},
    \citeyear{groenendijk-questions}; \citealp{groenendijk-logic}}
    provide a logically attractive theory of the semantics of natural
    language questions, commonly referred to as the \emph{partition
    theory}.  Two central notions in this theory are \emph{entailment
    between questions} and \emph{answerhood}.  For example, the question
    \SHANphrase{Who is going to the party?} entails the question \SHANphrase{Is
    John going to the party?}, and \SHANphrase{John is going to the party}
    counts as an answer to
    both.  Groenendijk and Stokhof define these two notions in terms of
    partitions of a set of possible worlds.

    We provide a syntactic characterization of \emph{entailment between
    questions} and \emph{answerhood}. We show that answers are, in some
    sense, exactly those formulas that are built up from instances of the
    question.  This result lets us compare the partition theory with other
    approaches to interrogation---both linguistic analyses, such as
    Hamblin's and Karttunen's semantics, and computational systems, such as
    Prolog. Our comparison separates a notion of answerhood into three aspects:
    \emph{equivalence} (when two questions or answers are interchangeable),
    \emph{atomic answers} (what instances of a question count as answers),
    and \emph{compound answers} (how answers compose).
\end{chapterabstract}

{\let\thefootnote\relax\footnotetext{We would like to thank Patrick Blackburn,
Paul Dekker, Jeroen Groenendijk, Maarten
Marx, Robert van Rooy, Stuart Shieber, and the anonymous referees
for their useful comments and discussions.
The 13th European Summer School in Logic, Language and Information, the
13th Amsterdam Colloquium, and
Stanford University's Center for the Study of Language and Information
provided stimulating environments that led to this
collaboration.
Part of the work presented here was carried out by the second author
during a visit to Nancy as part of the INRIA funded partnership between 
LIT (Language and Inference Technology, ILLC, University of 
Amsterdam) and LED (Langue et Dialogue, LORIA, Nancy).
The first author is supported by the United States
National Science Foundation under Grant IRI-9712068.}}

\section{The partition theory of questions}
\label{s:partition}

An elegant account of the semantics of natural language
questions from a logical and mathematical perspective is the one provided
by \citet{groenendijk-studies}.  According to
them, a question denotes a partition of
a logical space of possibilities. In this section, we give a brief summary
of this influential theory, using a notation slightly different from
\SHANcitegenitive{\citet{groenendijk-logic}} presentation.

A question is essentially a first order formula, possibly with free
variables. We will denote a question by
$\SHANq\phi$, where $\phi$ is a first order formula.
(We will also denote a set of questions by $\SHANq\Phi$, where $\Phi$
is a set of first order formulas.)
An answer is also a first order formula, but one that stands in a
certain \emph{answerhood} relation with respect to the question,
spelled out later in this section.  For example,
the statement \SHANphrase{Everyone is going to the party} ($\forall x Px$)
will turn out to answer the question \SHANphrase{Who is going to the
party?} ($\SHANq Px$).

We assume that equality is in the language, so one can ask questions
such as \SHANphrase{Who is John?} ($\SHANq x\SHANapprox j$).
We also assume
that, for every function symbol---including constants---it is indicated
whether it is interpreted rigidly or not.
Intuitively, for a function symbol to be rigid means that its denotation
is known. For example, under the notion of answerhood that we will
introduce below, it is only appropriate to answer
\SHANphrase{Who is going to the party?} ($\SHANq Px$) with \SHANphrase{John is going to the party}
($Pj$) if it is known who \SHANphrase{John} is---in other words, if $j$ is rigid.
Also, for \SHANphrase{Who is John?} ($\SHANq x\SHANapprox j$) to be a non-trivial question, \SHANphrase{John}
must have a non-rigid interpretation.

Questions are interpreted relative to first order modal structures
with constant domain. That is, a model is of the form $(W,D,I)$, where
$W$ is a set of worlds, $D$ is a domain of entities, and $I$ is an
interpretation function assigning extensions to the predicates and
function symbols, relative to each world.
Furthermore, we only consider models that give rigid function
symbols the same extension in every world.
Relative to such a model $M=(W,D,I)$, a question~$\SHANq\phi$ expresses
a partition of~$W$, in other words an equivalence relation over~$W$:
\begin{equation} \label{e:partition}
[\SHANq\phi]\sb M = \{\, (w,v)\in W\sp2 \mid \forall g \colon
  M,w,g\models\phi \Leftrightarrow M,v,g\models\phi \,\}
\enspace.
\end{equation}
Roughly speaking, two worlds are equivalent if one cannot tell
them apart by means of the question~$\SHANq\phi$, in other words if
$\phi$ is never true in one world and false in the other.
In general, any set of questions~$\SHANq\Phi$
also expresses a partition of~$W$, namely the intersection of the
partitions expressed by its elements:
\begin{equation}
\begin{split}
[\SHANq\Phi]\sb M &= \textstyle\bigcap\nolimits\sb{\phi\in\Phi} [\SHANq\phi]\sb M \\
          &= \{\, (w,v)\in W\sp2 \mid \forall \phi\in\Phi\colon \forall g\colon
             M,w,g\models\phi \Leftrightarrow M,v,g\models\phi \,\}
\enspace.
\end{split}
\end{equation}
Entailment between questions is defined as a \emph{refinement} relation
among partitions (i.e.,\ equivalence relations): An equivalence relation~$A$
is a subset of another equivalence relation~$B$ if every equivalence class
of~$A$ is contained in a class of~$B$.
\begin{equation}
\SHANq\Phi\models \SHANq\psi \quad\text{iff}\quad
\forall M \colon [\SHANq\Phi]\sb M \subseteq [\SHANq\psi]\sb M
\enspace.
\end{equation}
The intuitive interpretation of $\SHANq\Phi\models\SHANq\psi$ is that resolving
the questions~$\SHANq\Phi$ implies (requires) resolving the question~$\SHANq\psi$.
In other words, the questions~$\SHANq\Phi$ distinguish between
more worlds than the question~$\SHANq\psi$ does.

A more fine-grained notion of entailment is as follows
\citep{groenendijk-logic}. Let $\chi$ be a first order formula
with no free variables, and let $M\models\chi$ mean that
$M,w\models\chi$ for all~$w$.
\begin{equation}
\label{e:entailment}
\SHANq\Phi\models\sb\chi \SHANq\psi \quad\text{iff}\quad \forall M\colon
M\models\chi \Rightarrow
[\SHANq\Phi]\sb M \subseteq [\SHANq\psi]\sb M
\enspace.
\end{equation}
Pronunciation: The questions $\SHANq\Phi$ entail the question $\SHANq\psi$
\emph{in the context of $\chi$} (or, \emph{given $\chi$}).
The context~$\chi$ is intended to capture assertions in the common ground:
If it is commonly known that everyone who got invited to the party is
going, and vice versa ($\forall x (Ix \SHANliff Px)$), then the questions \SHANphrase{Who
got invited?} ($\SHANq Ix$) and \SHANphrase{Who is going?} ($\SHANq Px$) entail each other.

This entailment relation between questions
allows us to define a notion of answerhood.
\begin{SHANdefinition}[Answerhood]
\label{def:answerhood}
Let $\SHANq\phi$ be a question and
$\psi$ a first order formula without free variables. We say that $\psi$ is an
\emph{answer} to $\SHANq\phi$ if $\SHANq\phi\models\SHANq\psi$.
\end{SHANdefinition}
\noindent
According to this notion (termed \emph{licensing} by
\citet{groenendijk-logic} and \emph{aboutness} by
\citet{lewis-relevant}), \SHANphrase{Everyone is going to the party}
($\forall x Px$) is an answer to \SHANphrase{Who is going to the
party?} ($\SHANq Px$), because $\SHANq Px \models \SHANq\forall x Px$.

Note that, under this definition, any contradiction or tautology counts as
an answer to any question.  \citet{groenendijk-studies}
define a stricter notion of \emph{being a partial semantic answer}
(\emph{pertinence} in \citealp{groenendijk-logic}),
which excludes these two trivial cases by formalizing
Grice's Maxims of Quality and Quantity, respectively.  In this paper, however,
we will stick to the simpler criterion of answerhood as defined
above, which can be seen to formalize Grice's Maxim of Relation. 
This is not because we believe the tautology and the contradiction to be
appropriate answers, but rather because they are
trivial cases that do not play a very interesting role in a theory of 
answerhood, merely complicating the picture by requiring a 
syncategorematic treatment. By the way, the inappropriateness of 
tautological and contradictory assertions is not specific to
the case of questions and answers.

We now have a semantic notion of answerhood, that of 
Definition~\ref{def:answerhood}, telling us what
counts as an answer to a question. For practical purposes, it
is useful to also characterize this notion \emph{syntactically}.
Can one give a simple syntactic property that is a necessary
and sufficient condition for answerhood? The next section shows how.

\section{A syntactic characterization of answerhood}
\label{s:syntactic}

First, let us look at a partial result discussed by
\citet{groenendijk-studies} and
\citet{kager-questions}. Define rigidity of
terms and formula instances in the following straightforward way.

\begin{SHANdefinition}[Rigidity]
\label{def:rigidity}
A term is \emph{rigid} if it is composed of variables and rigid
  function symbols.
A formula~$\phi$ is a \emph{rigid instance} of another formula~$\psi$
  if $\phi$ can be obtained from
  $\psi$ by uniformly substituting rigid terms for variables.
An identity statement $s\SHANapprox t$ is \emph{rigid} if the terms $s$ and $t$
  are rigid.
\end{SHANdefinition}
\noindent
For example, if $c$ is a rigid constant,
then rigid instances of $Px$ include $Pc$ and $Px$. The
identity statement $c\SHANapprox x$ is also rigid. Notice that rigid instances
are not necessarily rigid: If $c$ is rigid, then $Rcd$ is a 
rigid instance of $Rxd$ even if the constant $d$ is not rigid. 

Groenendijk and Stokhof and Kager observed that rigid
instances of a question constitute answers to that question. By a
simple inductive argument, one can generalize this a bit.

\begin{SHANdefinition}[Development]
\label{def:development}
A formula~$\psi$ is a \emph{development} of another
  formula~$\phi$ (written $\phi\leq \psi$) if $\psi$ is built up from
rigid instances of $\phi$ and rigid identity statements using
boolean connectives and quantifiers.
In other words, the developments of~$\phi$ are the formulas~$\psi$
generated by
\begin{equation}
\psi ::= \phi\sp\sigma \mid t\sb1\SHANapprox t\sb2 \mid \neg\psi \mid
\psi\sb1\land\psi\sb2 \mid \psi\sb1\lor\psi\sb2 \mid \exists x \psi
\mid \forall x \psi \enspace,
\end{equation}
where
$t\sb1$ and $t\sb2$ are rigid terms, and $\sigma$ substitutes rigid terms
for variables. 
For example, if $c$ and~$d$ are rigid constants, then both $Pc\land Pd$ and
$\exists x \bigl(Px\land \neg(x\SHANapprox c)\bigr)$ are developments of $Px$.
\end{SHANdefinition}

\begin{SHANtheorem}\label{thm:gsk}If $\phi\leq \psi$ then $\SHANq\phi\models \SHANq\psi$.
\end{SHANtheorem}

\begin{proof}
By induction on the size of $\phi$.
\end{proof}

Theorem~\ref{thm:gsk} says that every development of~$\phi$ is an answer
to~$\SHANq\phi$.
We can prove a converse of sorts: Every answer to~$\SHANq\phi$ is
\emph{equivalent} to a development of~$\phi$.
The proof proceeds roughly as follows. Recall that answerhood is defined
in terms of question entailment. Suppose that $\psi$ is an answer
to the question $\SHANq\phi$. Then $\SHANq\phi\models \SHANq\psi$. Using
a certain translation procedure, we can turn this question entailment
into an ordinary first order entailment 
$\SHANq\phi\sp\#\models \SHANq\psi\sp\#$. Next, 
we apply Craig's interpolation theorem for first order logic to 
obtain an interpolant~$\vartheta$, from which we can recover an
answer to the original question~$\SHANq\phi$. Finally, we show that this
answer is in fact equivalent to~$\psi$, and a development of~$\phi$. 

Before we present the theorem and its proof in more detail, we will
mention the procedure that we use to translate question entailment
into first order entailment. For any first order formula $\phi$, let
$\phi\sp*$ be the result of priming all non-rigid non-logical symbols.
For example, if $c$ is a rigid constant symbol and $d$ is a non-rigid
constant symbol, and
\begin{equation}
    \phi = \exists x(Pxc\land\neg Pyd) \enspace,
\end{equation}
then 
\begin{equation}
    \phi\sp* = \exists x(P'xc\land\neg P'yd') \enspace.
\end{equation}
Next, translate any question $\SHANq\phi$ with
free variables $x\sb 1,\ldots,x\sb n$ to the formula
\begin{equation}
    \SHANq\phi\sp\# = \forall x\sb 1\ldots\forall x\sb n(\phi\leftrightarrow\phi\sp*)
    \enspace.
\end{equation}
For instance, the example question~$\SHANq\phi$ translates to
\begin{equation}
    \SHANq\phi\sp\# = \forall y \bigl(\exists x(Pxc\land\neg Pyd) \leftrightarrow
                         \exists x(P'xc\land\neg P'yd') \bigr)\enspace.
\end{equation}
Then the question entailment
$\SHANq\phi\models\sb\chi \SHANq\psi$ is valid iff the first order
entailment
$\SHANq\phi\sp\#,\chi,\chi\sp*\models\SHANq\psi\sp\#$ is valid. 
A detailed proof is given elsewhere \citep{cate-p2p}, based 
a one-to-one correspondence
between first order models for the enriched signature (with primed 
copies of the non-rigid non-logical symbols) and first order modal 
structures for the original signature 
\emph{that contain exactly two worlds}.

Using this translation, we can now formulate and prove the following
converse-modulo-equivalence of Theorem~\ref{thm:gsk}. 

\begin{SHANtheorem}\label{thm:syntactic}
Suppose that $\SHANq\phi\models\sb\chi \SHANq\psi$, and let $\vec{y}$ be the 
free variables of $\psi$.
Then there exists some
formula~$\vartheta$
with no free variables beside $\vec{y}$
such that $\phi\leq \vartheta$ and $\chi \models \forall\vec{y}(\psi
\leftrightarrow\vartheta)$.
\end{SHANtheorem}

\begin{proof}
First, we will prove the special case where $\phi$ is an atomic formula,
say~$P\vec{x}$.  Suppose that $\SHANq P\vec{x} \models\sb\chi \SHANq\psi(\vec{y})$.
Then, using the translation procedure discussed above,
\begin{equation}
\forall\vec{x}(P\vec{x}\leftrightarrow P'\vec{x}),\chi,\chi\sp* \models
\forall\vec{y}(\psi(\vec{y})\leftrightarrow\psi\sp*(\vec{y}))
\enspace.
\end{equation}
As a fact of first order logic, we can replace the universally
quantified variables in the consequent by some freshly chosen constants $\vec{c}$.
This results in\looseness=-1\relax
\begin{equation}
\forall\vec{x}(P\vec{x}\leftrightarrow P'\vec{x}),\chi,\chi\sp* \models
\psi(\vec{c})\leftrightarrow\psi\sp*(\vec{c})
\enspace,
\end{equation}
and, from this,
\begin{equation}
\label{e:interpolatee}
\forall\vec{x}(P\vec{x}\leftrightarrow P'\vec{x}), \chi\sp*, \psi\sp*(\vec{c})
\models \chi\to\psi(\vec{c})
\enspace.
\end{equation}
By Craig's interpolation theorem for first order logic, we can
construct an interpolant $\vartheta(\vec{c})$ such that
\begin{gather}
\label{e:interpolation-1}
\forall\vec{x}(P\vec{x}\leftrightarrow P'\vec{x}), \chi\sp*, \psi\sp*(\vec{c})
    \models \vartheta(\vec{c}) \enspace,\\
\label{e:interpolation-2}
\vartheta(\vec{c}) \models \chi\to\psi(\vec{c}) \enspace,
\end{gather}
and the only non-logical symbols in $\vartheta(\vec{c})$ are those
occurring on both sides of~\eqref{e:interpolatee}.
From the way the translation procedure $(\cdot)\sp*$ is set up, it follows
that the only non-logical symbols that $\chi\sp*$ and~$\psi\sp*$ on the one hand,
and $\chi$ and~$\psi$ on the other hand, have in common, are rigid
function symbols.  Thus, $\vartheta(\vec{c})$ contains no
non-logical symbols beside $P$, $\vec{c}$, and rigid function symbols.

Removing primes uniformly from all predicate and function symbols in
 \eqref{e:interpolation-1}, we get
 $\chi\models\psi(\vec{c})\to\vartheta(\vec{c})$.
From \eqref{e:interpolation-2}, we get the converse:
 $\chi\models\vartheta(\vec{c})\to\psi(\vec{c})$. Together, this gives us
\begin{equation}
    \chi\models\psi(\vec{c})\leftrightarrow \vartheta(\vec{c})
    \enspace.
\end{equation}
Since the constants $\vec{c}$ do not occur in $\chi$,
we can replace them by universally quantified variables.
This results in
\begin{equation}
    \chi\models \forall \vec{y} (\psi(\vec{y})\leftrightarrow
    \vartheta(\vec{y}))
    \enspace.
\end{equation}
Furthermore, $\vartheta(\vec{y})$ contains no non-logical symbols
beside $P$ and rigid function symbols. This means that
$\vartheta(\vec{y})$ is built up from instances of $P\vec{x}$,
rigid identity statements, $\top$, and $\bot$ using the boolean
connectives and quantifiers. As a last step, we replace $\top$ by 
$\forall\vec{x}P\vec{x}\lor\neg\forall\vec{x}P\vec{x}$ and $\bot$
by $\forall\vec{x}P\vec{x}\land\neg\forall\vec{x}P\vec{x}$. The
result is a development of $P\vec{x}$.

As for the general case, suppose $\SHANq\phi(\vec{x})\models\sb\chi \SHANq\psi$.
Choose a fresh predicate symbol~$P$ with the same arity as the number
of free variables of $\phi$. Then it follows that
$\SHANq P\vec{x} \models\sb{\chi\land\forall\vec{x}(P\vec{x}\SHANliff
  \phi(\vec{x}))} \SHANq\psi$. Apply the above strategy to obtain a development
$\vartheta$ of $\SHANq P\vec{x}$ such that
$\chi\land\forall\vec{x}(P\vec{x}\leftrightarrow
\phi(\vec{x}))\models \forall\vec{y}(\vartheta\leftrightarrow \psi)$.
Let $\vartheta'$ be the result of replacing all subformulas in
$\vartheta$ of the form $P\vec{z}$ by $\phi(\vec{z})$. Then
$\vartheta'$ is a development of $\phi$, and
$\chi\models\forall\vec{y}(\psi\leftrightarrow \vartheta')$.
\end{proof}
\noindent
Thus, the syntactic notion of development corresponds precisely to the
semantic notion of entailment between questions.

As an example, suppose that it is commonly known that everyone who got
invited to the party is going, and vice versa ($\chi=\forall x (Ix \SHANliff
Px)$), and the identity of John is known ($j$ is rigid).  In response to
the question \SHANphrase{Who is going to the party?} ($\SHANq\phi=\SHANq Px$),
it is appropriate to answer \SHANphrase{John is invited} ($\psi=Ij$).  As
assured by Theorem~\ref{thm:syntactic}, the answer~$\psi$ is equivalent to some
development~$\vartheta$ of~$\phi$ given~$\chi$; in this example, we can
take $\vartheta=Pj$.

Recall that the formula~$\chi$
represents an assertion in the common ground.
If there is no assertion in the common ground (i.e., $\chi=\top$),
then Theorems \ref{thm:gsk} and~\ref{thm:syntactic} reduce to the
following syntactic characterization of answerhood.

\begin{SHANcorollary}\label{cor:answerhood}
A formula~$\psi$ without free variables is an answer to~$\SHANq\phi$ iff $\psi$ is
  equivalent to a development of~$\phi$.
\end{SHANcorollary}

This result easily generalizes from entailment by a single question to
entailment by a set of questions.  (A development of a \emph{set} of
formulas~$\Phi$ is a formula built up from rigid instances of
\emph{elements} of~$\Phi$ and rigid identity statements using boolean
connectives and quantifiers) Likewise, if we
eliminate equality from the logical language, then the characterization
still applies.  (The appropriate notion of development is then
a formula built up from rigid instances using boolean connectives and
quantifiers)

This syntactic characterization is useful for at least two purposes. First, 
it makes possible a thorough investigation of the predictions
made by Groenendijk and Stokhof's theory of answerhood: It shows
what their semantic theory really amounts to, syntactically speaking.
Second, this result opens the way to practical question answering
algorithms: A question answering system can operate purely by
symbolic manipulation without referring to the semantics. The
remainder of this paper will concentrate on the first application: We will
use our characterization to clarify the relation between the partition
theory and other theories on the semantics of questions. 

\section{Comparing theories of questions}
\label{s:comparing}

So far, we have provided a syntactic characterization of the partition
theory notion of answerhood.  Our characterization affords us a new
perspective on exactly what assumptions are made by the partition theory,
because from its statement we can directly read off three aspects of
a notion of answerhood:
\begin{description}
\item [Equivalence] When are two questions or answers interchangeable?
      The partition theory considers formulas \emph{modulo logical equivalence}.
\item [Atomic answers] What instances of a question count as answers?
      The partition theory admits \emph{rigid} instances of the question as
      answers.
\item [Compound answers] How do answers compose?  The partition theory
      builds up answers \emph{using boolean connectives and quantifiers}.
\end{description}
We can now examine each aspect of answerhood in turn, and compare the
partition theory (in particular its notion of licensing) with other 
theories of answerhood.

\subsection{Equivalence}
\label{sec:equivalence}

In response to the question \SHANphrase{Is John friendly?}, all theories agree
that \eqref{e:john-is-friendly} counts as an answer and
\eqref{e:it-is-raining} does not (or only in a very specific context).  
But what about the responses in~\eqref{e:john-is-friendly-alt}?
\begin{exe}
\ex
\begin{xlist}
    \ex \label{e:john-is-friendly}John is friendly.
    \ex \label{e:it-is-raining}It is raining.
\end{xlist}
\ex \label{e:john-is-friendly-alt}
\begin{xlist}
    \ex \label{e:john-is-friendly-dir}John is not unfriendly.
    \ex \label{e:john-is-friendly-eqv}If $2+2=4$ then John is friendly,
        otherwise it is raining.
\end{xlist}
\end{exe}
Theories that base meanings on propositions as sets of possible worlds
equate the logically equivalent sentences \eqref{e:john-is-friendly},
\eqref{e:john-is-friendly-dir}, and~\eqref{e:john-is-friendly-eqv}, so they
predict that the three sentences are all acceptable answers.  These
theories include \SHANcitegenitive{\citet{hamblin-questions}} and
\SHANcitegenitive{\citet{karttunen-syntax}} proposals, where each question
denotes a set of answers, as well as \citeauthor{groenendijk-studies}'s
partition theory.  By contrast, theories such as that of
\citet{higginbotham-questions}, where meanings are based on the syntactic
form of formulas, treat the same three sentences as distinct and unrelated,
thus predicting only the
answer~\eqref{e:john-is-friendly}. As \cite{higginbotham-questions}
phrase it themselves: 
\begin{quote}\small
On the view of questions developed here they are individuated by their 
linguistic form, and count as distinct even if their content is in
some sense the same. This identity criterion contrasts with that
emerging most naturally from possible worlds semantics as applied to
questions, in which they would be individuated by their propositional
content; see, for example, Karttunen (1977). An immediate consequence
of this latter approach is that the question whether $p$ and the
question whether $q$ are identical if $p$ and $q$ are necessarily
equivalent. Similarly, \SHANphrase{What is the sum of 16 and 2?} and
\SHANphrase{What is the product of 3 and 6?} will express the same
question. This consequence does not follow for the view sketched here;
it remains for us an explicative task to analyse the relation
obtaining between questions when they are equivalent propositionally,
but not notationally. 
\end{quote}
Thus, if we want to rule in \eqref{e:john-is-friendly}
and~\eqref{e:john-is-friendly-dir} as answers yet rule out
\eqref{e:john-is-friendly-eqv}, then propositional equivalence faces the
problem of logical omniscience,
and notational equivalence faces the problem of syntactic variation.

How the partition theory handles equivalence is clear in the statement of
Corollary~\ref{cor:answerhood}, which bestows answerhood on not just
developments of the question but also all logically equivalent formulas.
We could change this definition of answerhood to use a finer-grained notion
of formula equivalence.  For instance, we might adopt some notion of
`strong equivalence' under which \eqref{e:john-is-friendly}
and~\eqref{e:john-is-friendly-dir} are equivalent to each other but not
to~\eqref{e:john-is-friendly-eqv}.  In this way, we can separate concerns
specific to answerhood from the general problem of logical omniscience, and
render the partition theory comparable to, say,
\citeauthor{higginbotham-questions}'s treatment of questions.
(Alternatively, one could invoke
Grice's Maxim of Manner to rule out answers such 
as~\eqref{e:john-is-friendly-eqv}.)

Equivalence depends on the \emph{context} in which questions and answers
occur.  For example, if it is known that Sue and Bill are John's parents,
then one can answer~\eqref{e:whose-mother-left} with~\eqref{e:sue-left}.
\begin{exe}
\ex
\begin{xlist}
    \ex \label{e:whose-mother-left}
        I know that Bill left, but did John's \textsc{mother} leave?
    \ex \label{e:sue-left}Sue left as well.
\end{xlist}
\end{exe}
The partition theory naturally handles such contexts: The
formula~$L(s)$ answers the question~$\SHANq L(m(j))$ given the
context~$\chi = s \SHANapprox m(j)$, in that the entailment $\SHANq L(m(j))
\models\sb\chi \SHANq L(s)$ is valid, as is easily seen through
Theorem~\ref{thm:syntactic}.  This example illustrates that what questions
and answers are equivalent depends on the common ground between the
questioner and the answerer.  This dependence is easier to capture in the
partition theory, which specifies an entailment relation between questions
and answers, than in theories where each question completely
determines its answers, as a set of propositions or closed formulas.

\subsection{Atomic answers}

Everybody agrees that the question \SHANphrase{Who left?} can be answered
with statements of the form \SHANphrase{[\dots] left}, but it is unclear what
noun phrases can fill the blank.
\begin{exe}
\ex
\begin{xlist}
    \ex \label{e:john-left}John left.
    \ex \label{e:unhelpful-left}The owner of US passport 126392058 left.
    \ex \label{e:unknown-left}John's mother left (but I'm not sure who she is).
\end{xlist}
\end{exe}
If the identity of John is known, then~\eqref{e:john-left} is usually an
acceptable answer, unlike~\eqref{e:unhelpful-left}, which is usually
unacceptable because the hearer is unlikely to know the denotation of the
noun phrase \SHANphrase{the owner of US passport 126392058}. 
Responses like~\eqref{e:unknown-left} fall in a gray area.
Note also that the question alone does not completely determine the
acceptable noun phrases: The question \SHANphrase{Who is the president of 
Mali?} calls for a different answer at a class exam than at a high
party \citep{aloni-quantification}.
\citet{ginzburg-resolving-1,ginzburg-resolving-2}
has raised important objections to simple-minded
theories of answerhood, by arguing that pragmatics
plays a role in determining what counts as an answer in
a given context. As he argues, not only does common ground knowledge
(whether the denotation of a term is known) play a role, but also the
goals and the intentions of the questioner. 

As shown by our syntactic characterization, the partition theory mandates
that the acceptable descriptions are precisely the rigid ones. That is,
only rigid instances of the question can be used to build
up an answer.  On this issue of acceptable descriptions, most
other semantic accounts of interrogation remain silent at best
\citep{hamblin-questions,karttunen-syntax,higginbotham-questions,krifka-structured-revised}.
Take for example Karttunen's theory, among the more precisely
specified of the alternatives.  It assigns denotations to interrogative
clauses by quantifying over individual concepts, not individuals.  In
principle, this move opens the door for non-rigid descriptions.  However,
perhaps since he was not conscious of the move, Karttunen
does not specify exactly which concepts to quantify over, in other words
exactly which descriptions to allow.  (To indiscriminately quantify over
all concepts would incorrectly predict that almost nobody knows the
full answer to any question.)\looseness=-1

Which terms are rigid depends on the context.
In general, as the common 
ground increases, more and more terms become rigid---the denotation of 
more and more terms becomes known.
Given that the appropriateness of answers depends on rigidity,
it is expected that what is
allowed as an answer depends on the amount of information that
is common ground. The partition theory nicely accounts for this, as 
we saw in Sect.\,\ref{sec:equivalence}. 

We view the rigidity criterion as an attempt to approximate whether a description
`specifies' an entity.  This approximation is not
perfect, as shown by the following exchange (provided to us by a referee).
\begin{exe}
\ex
{}[The Amsterdam Marathon course has been altered.]
\begin{xlist}
\exi{A:}Who will these changes affect most?
\exi{B:}The winner. He will be annoyed to have to run the last kilometer on
        cobblestones.
\end{xlist}
\end{exe}
B's answer above is perfectly acceptable even before the marathon
takes place and the winner becomes known (rigid).  Such examples,
together with
\SHANcitegenitive{\citet{ginzburg-resolving-1,ginzburg-resolving-2}}
observations, motivate
us to refine how our definition of answerhood models specification.
\citet{aloni-quantification}
has taken up this point by extending
the partition theory with the notion of \emph{conceptual covers}
to allow for much more fine-grained control over
what instances count as atomic answers.
Her theory also accounts for questions such as 
\SHANphrase{Who is who?}.

So far, we have not discussed another kind of atomic answers predicted by
the partition theory, namely (rigid) identity statements.  Identity
statements by themselves are usually trivial and infelicitous as answers,
but they are essential in compound answers such as \SHANphrase{Only John is
coming to the party} ($\forall x(Px \rightarrow x\SHANapprox j)$). 
Perhaps the definition of development could be improved by excluding identity
statements of the form $x\SHANapprox y$, where both $x$ and~$y$ are
variables. This would exclude the problematic answer $\exists x \exists y \neg
(x\SHANapprox y)$ (``there exist at least two entities'').

\subsection{Compound answers}
\label{s:compound}

All theories agree that the conjunction of two answers still counts as an
answer (if the conjuncts are mutually consistent).  For
example, because \SHANphrase{John left} and \SHANphrase{Mary left} are
both answers to \SHANphrase{Who left?}, the conjunction 
\SHANphrase{John left and Mary left} is again an answer.  Oftentimes
\citep[as in][]{hamblin-questions,karttunen-syntax,higginbotham-questions},
when questions are analyzed as
denoting a set of answers, the set is actually construed to contain the
\emph{atomic answers} only: the proposition that John left and the
proposition that Mary left, say, but not the proposition that John left and
Mary left.  It is then explicitly stated or implicitly understood that, not
only are elements of this set appropriate answers, but their conjunctions
are as well.

The partition theory takes rigid instances of the question to be atomic answers.
However, it goes far beyond licensing
conjunctions of
atomic answers: As Corollary~\ref{cor:answerhood} makes clear, it
allows arbitrary propositions composed from rigid instances using
boolean connectives and quantifiers.  Consequently, it predicts that all
of the following are appropriate answers.
\begin{exe}
\ex \label{e:developments}
\begin{xlist}
    \ex John left.
    \ex John left and Mary left.
    \ex \label{e:developments-disjunction}
        Either John left, or Mary left.
    \ex John did not leave.
    \ex If anyone left, then John did.
    \ex Everyone left.
    \ex Nobody left except John.
    \ex Somebody left.
    \ex Somebody other than John didn't leave.
\end{xlist}
\end{exe}
The view that developments approximate the logical compositionality of
answers thus prompts us to refine how our definition of answerhood models
composition.

On one hand, the boolean connectives illustrated
in~(\ref{e:developments}b\nobreakdash--d) are easy to deal with in a theory
of interrogation where the meaning of each question determines a set of
atomic answers.  For instance, if we wish to permit disjunctions
like~\eqref{e:developments-disjunction} as answers, we can simply let the
set of appropriate answers be the closure of the set of atomic answers
under disjunction in addition to conjunction.

On the other hand, quantifiers as illustrated
in~(\ref{e:developments}e\nobreakdash--i) are harder to treat in any theory
where (atomic) answers are closed formulas:
To build up the proposition that everyone left ($\forall x Lx$), we cannot
just combine formulas expressing that specific individuals left ($Lj$, $Lm$,
\dots).  That free variables are necessary is reflected in our definition
of rigid terms, which includes variables.  If we wish to treat questions as
sets of atomic answers while maintaining closure of answerhood under
quantification, one would have to introduce atomic answers that are not
propositional, in other words semantic objects that correspond
to logical formulas with free variables.

A move in the direction of non-propositional atomic answers has
been made by the so-called functional or ``structured meaning'' approach
to interrogation.  Under this approach, which has a long history,
``question meanings are functions that, when applied to the
meaning of the [short] answer, yield a proposition''
\cite[a recent exposition]{krifka-structured-revised}.
For instance, the meaning of
\SHANphrase{Who left?} is
\begin{equation}
\label{e:who-left}
    \SHANfun{x}Lx\enspace,
\end{equation}
which when applied to \SHANphrase{John} yields the proposition
\SHANphrase{John left}, as in~(\ref{e:developments}a).  To generate
compound answers, this approach can be turned on its head: by
viewing the \emph{responsive} phrase as the operator, as
\citet{ginzburg-interrogatives} puts it, the meaning
of \SHANphrase{everyone} becomes the functional $\SHANfun{c}\forall x\,c(x)$,
which when applied to~\eqref{e:who-left} yields the proposition
$\forall{x}Lx$, as in~(\ref{e:developments}f).  In this way,
generalized quantifiers can form compound answers as well as atomic ones.

Incidentally, notice that the answers
in~(\ref{e:developments}a\nobreakdash--g) seem more appropriate than those
in~(\ref{e:developments}h\nobreakdash--i).  These two groups of answers are
distinguished by the following criterion: On one hand, the answers 
in~(\ref{e:developments}a\nobreakdash--g) correspond to formulas built
up from atomic answers and negations thereof using 
conjunction, disjunction, and universal quantification (but not existential
quantification). On the other hand, the answers in~(\ref{e:developments}h\nobreakdash--i)
are not of this type, as they essentially involve
existential quantification. A simple modification of the
definition of development allows us to capture these observations.

\section{Conclusion}

We presented a syntactic characterization of a notion of answerhood for the
partition semantics of questions.  In terms of the partition theory itself,
this result explains the meaning of a question in terms of the form of its
answers.  Moreover, in relation to other theories of interrogation, this
result lets us separate a notion of answerhood into three aspects:
\begin{enumerate}
\item equivalence (when two questions or answers are interchangeable),
\item atomic answers (what instances of a question count as answers), and
\item compound answers (how answers compose). 
\end{enumerate}
Organized along these three dimensions, theories of questions can then be
compared in a principled way.  In particular, theories differ in their
notions of compound answers, and hence in how they hypothesize answers
compose.

Apart from a more detailed comparison of theories of questions, we are
investigating two directions for further research. First, as 
we mentioned already in Sect.\,\ref{s:syntactic}, our result bears on
computational question answering algorithms. For example, we conceive
of Prolog as a question answering algorithm in the partition theory
sense \citep{cate-p2p}.

Second, we suggested several ways in which the definition of
development might be adjusted to better reflect our intuitions concerning what
counts as an answer to a question. These adjustments, like our original
definition of development, were stated syntactically, but we naturally wonder
whether they correspond to sensible variants of the (model-theoretic)
semantics given in~(\ref{e:partition}). To illustrate briefly,
consider again the alternative definition 
of developments suggested at the end of the previous section, according to
which compound answers must be built up from atomic answers without 
existential quantification. A well-known model-theoretic
result states that the first order properties that can be 
expressed without existential quantification are precisely the 
ones that are preserved under taking generated submodels. 
We leave it as an open question whether this model-theoretic
requirement can be given an intuitive linguistic motivation. 
Perhaps this also explains why
\SHANphrase{John left} is intuitively a good answer to the
question \SHANphrase{Did anybody leave?}.

\bibliographystyle{chicago}
\bibliography{shan}

\end{document}